\documentclass[final,12pt]{colt2021} 

\title[Online Confidence Intervals for RKHS Elements]{Open Problem: Tight Online Confidence Intervals for RKHS Elements }
\usepackage{times}

\coltauthor{%
 \Name{Sattar Vakili} \Email{sattar.vakili@mtkresearch.com}\\
 \addr MediaTek Research, UK
 \AND
 \Name{Jonathan Scarlett} \Email{scarlett@comp.nus.edu.sg}\\
 \addr National University of Singapore%
 \AND
 \Name{Tara Javidi} \Email{tjavidi@eng.ucsd.edu }\\
 \addr University of California, San Diego%
}

\usepackage{setspace}

\usepackage[utf8]{inputenc} 
\usepackage[T1]{fontenc}    
\usepackage{hyperref}  
\hypersetup{
    colorlinks=true,
    linkcolor=blue,
    citecolor =blue,
    filecolor=magenta,      
    urlcolor=magenta,
}
\usepackage{url}            
\usepackage{booktabs}       
\usepackage{amsfonts}       
\usepackage{nicefrac}       
\usepackage{microtype}      
\usepackage{lipsum}
\usepackage{graphicx}

\usepackage{amsmath}
\usepackage{amsfonts}
\usepackage{bm}
\usepackage{amssymb}
\usepackage{color}
\usepackage{xcolor}
\usepackage{hyperref}
\usepackage{amsmath, amssymb, graphicx, url}

\usepackage{cleveref}
\crefformat{section}{\S#2#1#3} 
\crefformat{subsection}{\S#2#1#3}
\crefformat{subsubsection}{\S#2#1#3}

\usepackage{mathrsfs}
\usepackage{makecell}

\def\x{\mathbf{x}}

\def\phib{\bm{\phi}}
\def\Ib{\bm{I}}

\def\Hc{\mathcal{H}}

\def\TP{\top}

\def\argmax{\text{argmax}}

\def\phib{\bm{\phi}}

\def\Kb{\mathbf{K}}
\def\kb{\mathbf{k}}

\def\yb{\mathbf{y}}
\def\Ib{\mathbf{I}}

\def\Lambdab{\mathbf{\Lambda}}
\def\wb{\mathbf{w}}
\def\zb{\mathbf{z}}

\def\Rr{\mathbb{R}}
\def\Nn{\mathbb{N}}
\def\E{\mathbb{E}}

\def\Xc{\mathcal{X}}
\def\Rc{\mathcal{R}}
\def\Oc{\mathcal{O}}
\def\Oct{\tilde{\mathcal{O}}}
\def\Fc{\mathcal{F}}

\newtheorem{assumption}{Assumption}

\begin{document}

\maketitle

\begin{abstract}%

Confidence intervals are a crucial building block in the analysis of various online learning problems. The analysis of kernel based bandit and reinforcement learning problems utilize confidence intervals applicable to the elements of a reproducing kernel Hilbert space (RKHS).  However, the existing confidence bounds do not appear to be tight, resulting in suboptimal regret bounds.  In fact, the existing regret bounds for several kernelized bandit algorithms (e.g., GP-UCB, GP-TS, and their variants) may fail to even be sublinear. It is unclear whether the suboptimal regret bound is a fundamental shortcoming of these algorithms or an artifact of the proof, and the main challenge seems to stem from the online (sequential) nature of the observation points. We formalize the question of online confidence intervals in the RKHS setting and overview the existing results. 

\end{abstract}


\vspace*{-1ex}
\section{Introduction}

The kernel trick provides an elegant and natural technique to extend linear models to non-linear models with a great representation power. In the past decade, numerous works have studied bandit and reinforcement learning problems under the assumption that the reward function conforms to a kernel-based model~\citep[]{srinivas2010gaussian,Krause11Contexual, wang2014theoreticalGPEI,   nguyen2017regretGPEI,  Scarlett2017Lower, Chowdhury2017bandit,wang2018meta,kandasamy2018parallelised,   Javidi, yang2020provably,shekhar2020multi,bogunovic2020corruption, zhou2020neuralUCB,Vakili2020Scalable,vakili2020information, Janz2020SlightImprov, cai2020lower, zhang2020neuralTS}.

The analysis of online learning problems with a kernel-based model typically utilizes confidence intervals applicable to the elements of a reproducing kernel Hilbert space (RKHS). However, the state-of-the-art confidence intervals in this setting~\citep{Chowdhury2017bandit} do not appear to be tight, resulting in suboptimal regret bounds.
The main challenge seems to stem from the online (sequential) nature of the observation points, in contrast to an offline (fixed in advance) design. 
We first overview the existing results, and then formalize the open problem of tight confidence intervals for the RKHS elements under the online setting. We also discuss the consequences of these bounds on the regret performance. 
For clarity of exposition, we focus on bandit problems and the GP-UCB algorithm~\citep{srinivas2010gaussian, Chowdhury2017bandit}, but the problem is equally relevant to reinforcement learning problems and other algorithms such as GP-TS. 

\vspace*{-1ex}
\section{Problem Setup}

Consider a positive definite kernel $k:\Xc \times \Xc\rightarrow \Rr$ with respect to a finite Borel measure, 
where $\Xc\subset \Rr^d$ is a compact
set. Let $\Hc_k$ denote the RKHS corresponding to $k$, defined as a Hilbert space equipped with an inner product $\langle.,.\rangle_{\Hc_k}$ satisfying the following: $k(.,x)\in \Hc_k$, $\forall x\in \Xc$, and $\langle f,k(.,x)\rangle_{\Hc_k}=f(x)$, $\forall x\in\Xc, \forall f \in \Hc_k$ (reproducing property). The typical assumption in kernel-based models is that the \emph{objective function} $f$ satisfies $f\in\Hc_k$ for a known kernel $k$. Let $\{\lambda_m\}_{m=1}^\infty$ and $\{\phi_m\}_{m=1}^\infty$ denote the Mercer eigenvalues and eigenfeatures of $k$, respectively~\citep[see, e.g.,][Theorem~$4.1$]{Kanagawa2018}. 
Using Mercer's representation theorem~\citep[see, e.g.,][Theorem~$4.2$]{Kanagawa2018}, an alternative representation for $f\in\Hc_k$ is given by
\begin{eqnarray}\label{kernelmodel}
f(x) = \wb^{\TP} \Lambdab^{\frac{1}{2}}\phib(x),
\end{eqnarray}
where $\wb = [w_1,w_2,...]^\TP$ and $\phib(x) = [\phi_2(x), \phi_2(x),...]^{\TP}$ are the (possibly infinite-dimensional) \emph{weight} and feature vectors, and $\Lambdab$ is a (possibly infinite dimensional) diagonal matrix with $\Lambdab_{i,j} = \lambda_i$, if $i=j$. The RKHS norm of $f$ satisfies $\|f\|_{\Hc_k} = \|\wb\|_{\ell^2}$.


\paragraph{Kernelized Bandits:}
Consider an online learning setting where a learning algorithm is allowed to collect a sequence of noisy observations $\{(x_i, y_i)\}_{i=1}^\infty$, where $y_i=f(x_i)+\epsilon_i$ with $\epsilon_i$ being well-behaved noise terms. The objective is to get as close as possible to the maximum of $f$. The performance of the algorithm is measured in terms of regret, defined as the cumulative loss in the values of the objective function at observation points, compared to a global maximum:
\begin{eqnarray}
\Rc(N) = \sum_{i=1}^N \left(f(x^*) - f(x_i)\right), 
\end{eqnarray}
where $x^*\in \argmax_{x\in \Xc} f(x)$ is a global maximum.  Under the assumption $f \in \Hc_k$, this setting is often referred to as that of kernelized bandits, Gaussian process (GP) bandits, or Bayesian optimization. The latter two terms are motivated by the algorithm design which often employs a GP surrogate model. Throughout this paper, we make the following assumptions.
\begin{assumption}\label{ass1}
The RKHS norm of $f$ is bounded as $\|f\|_{\Hc_k}\le B$, for some $B>0$. Moreover, the noise terms are i.i.d. sub-Gaussian random variables, i.e., for some $R>0$, $\E[\exp(\eta\epsilon_i)]\le \exp(\frac{\eta^2R^2}{2})$, $\forall \eta\in \Rr, \forall i \in \Nn$. 
\end{assumption}

In online learning problems, the observation points are collected sequentially. In particular, the observation point $x_{i+1}$ is determined after all the values $\{(x_j, y_j)\}_{j=1}^i$ are revealed. This is in contrast to an offline design, where the data points are fixed in advance. We next formalize this distinction.
\begin{definition}
i) In the \textbf{online setting}, for the sigma algebras $\Fc_i = \sigma(x_1,x_2,\dots, x_{i+1}, \epsilon_1, \epsilon_2, \dots, \epsilon_i)$, $i\ge 1$, it holds that $x_i$ and $\epsilon_i$ are $\Fc_{i-1}$ and $\Fc_i$ measurable, respectively.
ii) In the \textbf{offline setting}, for all $i\ge 1$, it holds that $x_i$ is independent of all $\epsilon_j$, $j \ge 1$.
\end{definition}

\paragraph{Surrogate GP Model:}

It is useful for algorithm design to employ a zero-mean surrogate GP model $\hat{f}$ with kernel $k$ which provides a surrogate posterior mean (regressor) and a surrogate posterior variance (uncertainty estimate) for the kernel-based model.  Defining $\mu_n(x) = \E\big[\hat{f}(x)|\{(x_i,y_i)\}_{i=1}^{n}\big]$ and $\sigma_n^2(x) = \E\big[(\hat{f}(x) - \mu_n(x))^2|\{(x_i,y_i)\}_{i=1}^{n}\big]$, it is well known that
$
\mu_n(x) =  \zb^{\TP}_n(x)\yb_n$ and 
$\sigma_n^2(x) = k(x,x) - \kb_n^{\TP}(x)  (\lambda^2 \Ib_n+\Kb_n)^{-1}\kb_n(x), 
$
where $\kb_n(x) = [k(x,x_1), k(x,x_2), \dots, k(x,x_n)]^{\TP}$, $\Kb_n$ is the positive definite kernel matrix $[\Kb_n]_{i,j} = k(x_i, x_j)$, $\zb_n(x) = (\lambda^2\Ib_n + \Kb_n)^{-1}\kb_n(x)$, $\Ib_n$ is the identity matrix of dimension $n$, and $\lambda>0$ is a regularization parameter.

\vspace*{-1ex}
\section{Confidence Intervals Applicable to RKHS Elements}

Deriving confidence intervals applicable to RKHS elements is significantly more challenging in the online setting compared to the offline setting. In the latter case, \citep{vakili21} showed that for any fixed $x\in \Xc$, we have with probability at least $1-\delta$ that
$
|f(x) - \mu_n(x)| \le \rho_0(\delta)\sigma_n(x),
$
where $\rho_0(\delta)=  B+ \frac{R}{\lambda} \sqrt{2\log(\frac{2}{\delta})}$, $B$ and $R$ are the parameters specified in Assumption~\ref{ass1}, and $\lambda$ is the regularization parameter of the surrogate GP model. Moreover, when $f$ is Lipschitz (or H\"older) continuous~\citep[that is true with typical kernels; see,][]{shekhar2020multi}, this easily extends to a uniform guarantee:  With probability at least $1-\delta$, we have uniformly in $x$ that $|f(x)-\mu_n(x)| = \Oc\big(
\big(B+\frac{R}{\lambda}\sqrt{d\log(n)+\log(\frac{1}{\delta})}\big)\sigma_n(x)
\big)$, where the implied constants in $\Oc(.)$ depend on the Lipschitz (or H\"older) continuity parameters.

In the online setting, strong uniform bounds are also well-known in the case of a linear model $f(x) = \wb^{\TP}x$: \cite{Abbasi2011} proved that, with probability $1-\delta$, uniformly over $x$,
\begin{eqnarray}\label{conf:lin}
|f(x) - \mu_n(x)| \le \rho_n(\delta)\sigma_n(x),
\end{eqnarray}
where $\rho_n(\delta)= B + \frac{R}{\lambda}\sqrt{d\log(\frac{1+n\bar{x}^2/\lambda^2}{\delta})} $ and $\bar{x}=\max_{x\in \Xc}\|x\|_{\ell^2}$.
The crux of the proof is a \emph{self-normalized bound for vector valued martingales} $S_n = \sum_{i=1}^n \epsilon_i x_i$~\citep[][Theorem $1$]{Abbasi2011},  which yields the following \emph{confidence ellipsoid} for $\wb$~\citep[][Theorem $2$]{Abbasi2011}: 
$
\|\wb-\hat{\wb}_n\|_{V_n} \le \lambda\rho_n(\delta), ~\text{with probability at least}~1-\delta
$, where $V_n= \lambda^2\Ib_d+\sum_{i=1}^nx_ix_i^{\TP}$.
This confidence ellipsoid for $\wb$ can then be represented in terms of the confidence interval for $f(x)$ given in~\eqref{conf:lin}.
Notice that the linear model is a special case of~\eqref{kernelmodel} with $\wb = [w_1,w_2,\dots, w_d]^{\TP}$ and $\phib(x) = x$ being $d$ dimensional weight and feature vectors respectively, and $\Lambda = \Ib_d$ being the square identity matrix of dimension $d$.

\cite{Chowdhury2017bandit} built on the self-normalized bound for the vector valued martingales to prove the following theorem for the kernel-based models. 
\begin{theorem}\label{TheoremOnK}
Under Assumption~\ref{ass1}, in the online setting, with probability at least $1-\delta$, we have for all $x\in \Xc$ that
\begin{eqnarray}
|f(x)-\mu_n(x)|\le \rho_n(\delta)\sigma_n(x),
\end{eqnarray}
where $\rho_n(\delta) = B+R\sqrt{2\left(\gamma_{n-1}+1+\log(\frac{1}{\delta})\right)}$, and $\gamma_n=\sup_{\{\x_i\}_{i=1}^n\subset\Xc}\log\det(\lambda^2\Ib_n+\Kb_n)$ is the maximal information gain at time $n$, which is closely related to the effective dimension associated with the kernel (e.g., see~\cite{srinivas2010gaussian, Valko2013kernelbandit}). 
\end{theorem}

Our open problem is concerned with improving this confidence interval. 

\paragraph{Open Problem.} Under Assumption~\ref{ass1}, in the online setting, consider the general problem of proving a confidence interval of the following form uniformly in $x\in \Xc$:
\begin{eqnarray}
|f(x)-\mu_n(x)|\le \rho_n(\delta)\sigma_n(x),~\text{with probability at least}~1-\delta.
\end{eqnarray}
What is the lowest growth rate of $\rho_n(\delta)$ with $n$? In particular, is it possible to reduce the confidence interval width in Theorem~\ref{TheoremOnK} by an $\Oct(\sqrt{\gamma_n})$ factor?

\vspace*{-1ex}
\section{Discussion}

Following standard UCB-based bandit algorithm techniques, it can be shown that the GP-UCB algorithm (namely, repeatedly choosing $x$ to maximize the current upper confidence bound) attains
\begin{eqnarray}
\Rc(N) = \Oct(\rho_N(\delta)\sqrt{N\gamma_N}), ~\text{with probability at least}~1-\delta.
\end{eqnarray}
Substituting $\rho_N(\delta)$ from Theorem~\ref{TheoremOnK}, we have $\Rc(N) = \Oct(\gamma_N\sqrt{N})$. Unfortunately, this is not always sublinear in $N$, since $\gamma_N$ can grow faster that $\sqrt{N}$, e.g., in the case of the Mat{\'e}rn family of kernels.  Hence, the regret bound can be trivial in many cases of interest. It is unknown whether this suboptimal regret bound is a fundamental shortcoming of GP-UCB or a result of suboptimal confidence intervals, but the latter appears likely to be the most significant factor. The same question can be asked about the analysis of many other bandit algorithms including GP-TS~\citep{Chowdhury2017bandit} and GP-EI~\citep{nguyen2017regretGPEI}, as well as KOVI in the reinforcement learning setting~\citep{yang2020provably}.

Comparing the results under the online and offline settings, we see a stark contrast of an $\Oc(\sqrt{\gamma_n})$ factor in the width of confidence intervals. We expect that the $\Oc(\sqrt{\gamma_n})$ factor in the confidence interval width in the online setting can be replaced by an $\Oct(d\log(n))$ term, resulting in an 
$\tilde{\Oc}(\sqrt{dN\gamma_N })$ regret bound. Roughly speaking, we are suggesting that a square root of the effective dimension of the kernel in the regret bound can be traded off for a square root of the input dimension. 

In a recent work addressing the above challenges and limitations, \cite{Janz2020SlightImprov} showed that partitioning the domain to many subdomains and fitting an independent GP to each one of them leads to an improved online confidence interval, based on local observations (see their Lemma~$5$). They leveraged this method to prove sublinear regret under Mat{\'e}rn kernels whenever the smoothness parameter $\nu$ exceeds one.  This is a significant improvement over the typical $\Oct(\gamma_N\sqrt{N})$ bound, which turns out to be sublinear only when $\nu > d/2$.

Of significant theoretical importance is a less practical algorithm \emph{SupKernelUCB}~\citep{Valko2013kernelbandit}, which achieves an $\tilde{\Oc}(\sqrt{N\gamma_N})$ regret bound for the kernelized bandit problem with a finite action set ($|\Xc|<\infty$). As noted in \cite{Janz2020SlightImprov}, the finite action set assumption can be relaxed to compact domains using a discretization argument; this turns out to contribute only an $\Oct(\sqrt{d\log(N)})$ factor to the regret bound~\citep[e.g., see,][for the details]{li2021gaussian}
. This bound is tight for the cases where a lower bound on regret is known, namely for commonly used squared exponential and Mat{\'e}rn kernels~\citep{Scarlett2017Lower, vakili2020information}.  

Very recently, a line of works has attained $\tilde{\Oc}(\sqrt{N\gamma_N})$ regret bounds for various other algorithms beyond SupKernelUCB.  However, all of them rely on more sophisticated techniques rather than a standard combination of GP-UCB and an online confidence bound.  Briefly, \cite{salgia2020} uses a tree-based partitioning on the domain along with localized search techniques, \cite{jamieson21} uses experimental design techniques to perform batch sampling and eliminate suboptimal inputs, and \cite{li2021gaussian} observes that the above-mentioned offline confidence bounds of \cite{vakili21} can be used within batches as long as the GP is reset at the start of each batch.

In view of this discussion, the improvement from $\tilde{\Oc}(\gamma_N\sqrt{N})$ to $\tilde{\Oc}(\sqrt{N\gamma_N})$ has now been obtained using several different algorithms, but it remains to determine whether GP-UCB (or a simple variation thereof) can achieve it.

\bibliography{bibliography}

\begin{thebibliography}{25}
\providecommand{\natexlab}[1]{#1}
\providecommand{\url}[1]{\texttt{#1}}
\expandafter\ifx\csname urlstyle\endcsname\relax
  \providecommand{\doi}[1]{doi: #1}\else
  \providecommand{\doi}{doi: \begingroup \urlstyle{rm}\Url}\fi

\bibitem[Abbasi-Yadkori et~al.(2011)Abbasi-Yadkori, P\'al, and
  Szepesv\'ari.]{Abbasi2011}
Yasin Abbasi-Yadkori, D\'avid P\'al, and Csaba Szepesv\'ari.
\newblock Improved algorithms for linear stochastic bandits.
\newblock In \emph{Advances in Neural Information Processing Systems}, pages
  2312--2320, 2011.

\bibitem[Bogunovic et~al.(2020)Bogunovic, Krause, and
  Scarlett]{bogunovic2020corruption}
Ilija Bogunovic, Andreas Krause, and Jonathan Scarlett.
\newblock Corruption-tolerant {Gaussian} process bandit optimization.
\newblock In \emph{International Conference on Artificial Intelligence and
  Statistics}, pages 1071--1081, 2020.

\bibitem[Cai and Scarlett(2021)]{cai2020lower}
Xu~Cai and Jonathan Scarlett.
\newblock On lower bounds for standard and robust {Gaussian} process bandit
  optimization.
\newblock In \emph{Proceedings of International Conference on Machine
  Learning}, 2021.

\bibitem[Camilleri et~al.(2021)Camilleri, Katz-Samuels, and
  Jamieson]{jamieson21}
Romain Camilleri, Julian Katz-Samuels, and Kevin Jamieson.
\newblock High-dimensional experimental design and kernel bandits.
\newblock In \emph{Int. Conf. Mach. Learn. (ICML)}, 2021.

\bibitem[Chowdhury and Gopalan(2017)]{Chowdhury2017bandit}
Sayak~Ray Chowdhury and Aditya Gopalan.
\newblock On kernelized multi-armed bandits.
\newblock In \emph{Proceedings of International Conference on Machine
  Learning}, pages 844--853, 2017.

\bibitem[Janz et~al.(2020)Janz, Burt, and Gonzalez]{Janz2020SlightImprov}
David Janz, David Burt, and Javier Gonzalez.
\newblock Bandit optimisation of functions in the matern kernel {RKHS}.
\newblock In \emph{Proceedings of Machine Learning Research}, volume 108, pages
  2486--2495. PMLR, 26--28 Aug 2020.

\bibitem[Javidi and Shekhar(2018)]{Javidi}
Tara Javidi and Shekhar Shekhar.
\newblock Gaussian process bandits with adaptive discretization.
\newblock \emph{Electronic Journal of Statistics}, 12\penalty0 (2):\penalty0
  3829--3874, 2018.

\bibitem[Kanagawa et~al.(2018)Kanagawa, Hennig, Sejdinovic, and
  Sriperumbudur]{Kanagawa2018}
Motonobu Kanagawa, Philipp Hennig, Dino Sejdinovic, and Bharath~K
  Sriperumbudur.
\newblock Gaussian processes and kernel methods: A review on connections and
  equivalences.
\newblock \emph{arXiv:1805.08845v1 [stat.ML]}, 2018.

\bibitem[Kandasamy et~al.(2018)Kandasamy, Krishnamurthy, Schneider, and
  P{\'o}czos]{kandasamy2018parallelised}
Kirthevasan Kandasamy, Akshay Krishnamurthy, Jeff Schneider, and Barnab{\'a}s
  P{\'o}czos.
\newblock Parallelised {Bayesian} optimisation via {Thompson} sampling.
\newblock In \emph{International Conference on Artificial Intelligence and
  Statistics}, pages 133--142, 2018.

\bibitem[Krause and Ong(2011)]{Krause11Contexual}
Andreas Krause and Cheng~S. Ong.
\newblock Contextual {G}aussian process bandit optimization.
\newblock In \emph{Advances in Neural Information Processing Systems 24}, pages
  2447--2455, 2011.

\bibitem[Li and Scarlett(2021)]{li2021gaussian}
Zihan Li and Jonathan Scarlett.
\newblock Gaussian process bandit optimization with few batches.
\newblock \emph{arXiv preprint arXiv:2110.07788}, 2021.

\bibitem[Nguyen et~al.(2017)Nguyen, Gupta, Rana, Li, and
  Venkatesh]{nguyen2017regretGPEI}
Vu~Nguyen, Sunil Gupta, Santu Rana, Cheng Li, and Svetha Venkatesh.
\newblock Regret for expected improvement over the best-observed value and
  stopping condition.
\newblock In \emph{Asian Conference on Machine Learning}, pages 279--294, 2017.

\bibitem[Salgia et~al.(2020)Salgia, Vakili, and Zhao]{salgia2020}
Sudeep Salgia, Sattar Vakili, and Qing Zhao.
\newblock A computationally efficient approach to black-box optimization using
  {G}aussian process models.
\newblock 2020.
\newblock https://arxiv.org/abs/2010.13997.

\bibitem[Scarlett et~al.(2017)Scarlett, Bogunovic, and
  Cevher]{Scarlett2017Lower}
Jonathan Scarlett, Ilija Bogunovic, and Volkan Cevher.
\newblock Lower bounds on regret for noisy {G}aussian process bandit
  optimization.
\newblock In \emph{Conference on Learning Theory}, pages 1723--1742, 2017.

\bibitem[Shekhar and Javidi(2020)]{shekhar2020multi}
Shubhanshu Shekhar and Tara Javidi.
\newblock Multi-scale zero-order optimization of smooth functions in an {RKHS}.
\newblock \emph{arXiv:2005.04832}, 2020.

\bibitem[Srinivas et~al.(2010)Srinivas, Krause, Kakade, and
  Seeger]{srinivas2010gaussian}
Niranjan Srinivas, Andreas Krause, Sham Kakade, and Matthias Seeger.
\newblock Gaussian process optimization in the bandit setting: no regret and
  experimental design.
\newblock In \emph{International Conference on Machine Learning}, pages
  1015--1022, 2010.

\bibitem[Vakili et~al.(2020)Vakili, Moss, Artemev, Dutordoir, and
  Picheny]{Vakili2020Scalable}
Sattar Vakili, Henry Moss, Artem Artemev, Vincent Dutordoir, and Victor
  Picheny.
\newblock Scalable {T}hompson sampling using sparse {G}aussian process models.
\newblock \emph{arXiv:2006.05356v3 [stat.ML]}, 2020.

\bibitem[Vakili et~al.(2021{\natexlab{a}})Vakili, Bouziani, Jalali, Bernacchia,
  and shan Shiu]{vakili21}
Sattar Vakili, Nacime Bouziani, Sepehr Jalali, Alberto Bernacchia, and Da~shan
  Shiu.
\newblock Optimal order simple regret for {G}aussian process bandits.
\newblock 2021{\natexlab{a}}.
\newblock https://arxiv.org/abs/2108.09262.

\bibitem[Vakili et~al.(2021{\natexlab{b}})Vakili, Khezeli, and
  Picheny]{vakili2020information}
Sattar Vakili, Kia Khezeli, and Victor Picheny.
\newblock On information gain and regret bounds in {Gaussian} process bandits.
\newblock In \emph{International Conference on Artificial Intelligence and
  Statistics}, pages 82--90, 2021{\natexlab{b}}.

\bibitem[Valko et~al.(2013)Valko, Korda, Munos, Flaounas, and
  Cristianini]{Valko2013kernelbandit}
Michal Valko, Nathan Korda, R\'{e}mi Munos, Ilias Flaounas, and Nello
  Cristianini.
\newblock Finite-time analysis of kernelised contextual bandits.
\newblock In \emph{Conference on Uncertainty in Artificial Intelligence}, pages
  654--663, 2013.

\bibitem[Wang et~al.(2018)Wang, Kim, and Kaelbling]{wang2018meta}
Zi~Wang, Beomjoon Kim, and Leslie~Pack Kaelbling.
\newblock Regret bounds for meta {Bayesian} optimization with an unknown
  {Gaussian} process prior.
\newblock In \emph{Advances in Neural Information Processing Systems}, pages
  10477--10488, 2018.

\bibitem[Wang and de~Freitas(2014)]{wang2014theoreticalGPEI}
Ziyu Wang and Nando de~Freitas.
\newblock Theoretical analysis of {Bayesian} optimisation with unknown
  {Gaussian} process hyper-parameters.
\newblock \emph{arXiv:1406.7758}, 2014.

\bibitem[Yang et~al.(2020)Yang, Jin, Wang, Wang, and Jordan]{yang2020provably}
Zhuoran Yang, Chi Jin, Zhaoran Wang, Mengdi Wang, and Michael Jordan.
\newblock Provably efficient reinforcement learning with kernel and neural
  function approximations.
\newblock In \emph{Advances in Neural Information Processing Systems},
  volume~33, 2020.

\bibitem[Zhang et~al.(2021)Zhang, Zhou, Li, and Gu]{zhang2020neuralTS}
Weitong Zhang, Dongruo Zhou, Lihong Li, and Quanquan Gu.
\newblock Neural {T}hompson sampling.
\newblock In \emph{International Conference on Learning Representations}, 2021.

\bibitem[Zhou et~al.(2020)Zhou, Li, and Gu]{zhou2020neuralUCB}
Dongruo Zhou, Lihong Li, and Quanquan Gu.
\newblock Neural contextual bandits with {UCB}-based exploration.
\newblock In \emph{International Conference on Machine Learning}, pages
  11492--11502, 2020.

\end{thebibliography}






\end{document}